# K-Histograms: An Efficient Clustering Algorithm for Categorical Dataset[*]

Zengyou He, Xiaofei Xu, Shengchun Deng, Bin Dong

*Department of Computer Science and Engineering Harbin Institute of Technology,*

*92 West Dazhi Street, P.O Box 315, P. R. China, 150001*

Email: zengyouhe@yahoo.com, xiaofei@hit.edu.cn, dsc@hit.edu.cn, db@hit.edu.cn

**Abstract** Clustering categorical data is an integral part of data mining and has attracted much attention recently. In this paper, we present *k*-histogram, a new efficient algorithm for clustering categorical data. The *k*-histogram algorithm extends the *k*-means algorithm to categorical domain by replacing the means of clusters with histograms, and dynamically updates histograms in the clustering process. Experimental results on real datasets show that *k*-histogram algorithm can produce better clustering results than *k*-modes algorithm, the one related with our work most closely.

## 1. Introduction

Clustering typically groups data into sets in such a way that the intra-cluster similarity is maximized while the inter-cluster similarity is minimized. The clustering technique has been extensively studied in many fields such as pattern recognition, customer segmentation, similarity search and trend analysis.

Most previous clustering algorithms focus on numerical data whose inherent geometric properties can be exploited naturally to define distance functions between data points. However, much of the data existed in the databases is categorical, where attribute values can't be naturally ordered as numerical values. An example of categorical attribute is *shape* whose values include *circle*, *rectangle*, *ellipse*, etc. Due to the special properties of categorical attributes, the clustering of categorical data seems more complicated than that of numerical data.

A few algorithms have been proposed in recent years for clustering categorical data [1~9]. In [5], the problem of clustering customer transactions in a market database is addressed. STIRR, an iterative algorithm based on non-linear dynamical systems is presented in [3]. The approach used in [3] can be mapped to a certain type of non-linear systems. If the dynamical system converges, the categorical databases can be clustered. Another recent research [4] shows that the known dynamical systems cannot guarantee convergence, and proposes a revised dynamical system in which convergence can be guaranteed.

In [6], the authors introduce a novel formalization of a cluster for categorical data by generalizing a definition of cluster for numerical data. A fast summarization based algorithm, CACTUS, is presented. CACTUS consists of three phases: *summarization*, *clustering*, and

[*] This work was supported by the High Technology Research and Development Program of China (No. 2002AA413310) and the IBM SUR Research Fund.

*validation*.

ROCK, an adaptation of an agglomerative hierarchical clustering algorithm, is introduced in [2]. This algorithm starts by assigning each tuple to a separated cluster, and then clusters are merged repeatedly according to the closeness between clusters. The closeness between clusters is defined as the sum of the number of "links" between all pairs of tuples, where the number of "links" is computed as the number of common neighbors between two tuples.

*Squeezer*, a one-pass algorithm is proposed in [1]. *Squeezer* repeatedly read tuples from dataset one by one. When the first tuple arrives, it forms a cluster alone. The consequent tuples are either put into existing clusters or rejected by all existing clusters to form a new cluster by given similarity function defined between tuple and cluster.

In [12], the authors propose the notion of *large item*. An item is *large* in a cluster of transactions if it is contained in a user specified fraction of transactions in that cluster. An allocation and refinement strategy, which has been adopted in partitioning algorithms such as k-means, is used to cluster transactions by minimizing the criteria function defined with the notion of large item.

K-modes, an algorithm extending the k-means paradigm to categorical domain is introduced in [7,9]. New dissimilarity measures to deal with categorical data is conducted to replace means with modes, and a frequency based method is used to update modes in the clustering process to minimize the clustering cost function.

In this paper, we present *k*-histogram, a new efficient algorithm for clustering categorical data. The *k*-histogram algorithm extends the *k*-means algorithm to categorical domain by replacing the means of clusters with histograms, and dynamically updates histograms in the clustering process. Compared to *k*-modes algorithm, we use the histogram data structure to describe a categorical data cluster instead of mode. Therefore, the *k*-modes algorithm is the one related with our work most closely. In the following parts, we will use the definitions and terminology presented in [7,9] to describe our algorithm.

## 2. Notations

Let $A_1, ..., A_m$ be a set of categorical attributes with domains $D_1, ..., D_m$ respectively. Let the dataset $D = \{X_1, X_2, ..., X_n\}$ be a set of objects described by $m$ categorical attributes, $A_1, ..., A_m$. The value set $V_i$ of $A_i$ is set of values of $A_i$ that are present in $D$. For each $v \in V_i$, the frequency $f(v)$, denoted as $f_v$, is number of objects $O \in X$ with $O.A_i = v$. Suppose the number of distinct attribute values of $A_i$ is $p_i$, we define the histogram of $A_i$ as the set of pairs: $h_i = \{(v_1, f_1), (v_2, f_2), ..., (v_{p_i}, f_{p_i})\}$. The histograms of the data set $D$ is defined as: $H = \{h_1, h_2, ..., h_m\}$.

Let $X, Y$ be two categorical objects described by $m$ categorical attributes. The dissimilarity measure between $X$ and $Y$ can be defined by the total mismatches of the corresponding attribute values of the two objects. The smaller the number of mismatches is, the more similar the two objects. Formally,

$$d_1(X,Y) = \sum_{j=1}^{m} \delta(x_j, y_j) \qquad (1)$$

where

$$\delta(x_j, y_j) = \begin{cases} 0 & (x_j = y_j) \\ 1 & (x_j \neq y_j) \end{cases} \quad (2)$$

Given the dataset $D = \{X_1, X_2, \ldots, X_n\}$ and an object $Y$, The dissimilarity measure between $X$ and $Y$ can be defined by the average of the sum of the distances between $X_i$ and $Y$.

$$d_2(D, Y) = \frac{\sum_{j=1}^{n} d_1(X_j, Y)}{n} \quad (3)$$

If we take the histograms $H = \{h_1, h_2, \ldots, h_m\}$ as the compact representation of the data set $D$, formula (3) can be redefined as (4).

$$d_3(H, Y) = \frac{\sum_{j=1}^{m} \phi(h_j, y_j)}{n} \quad (4)$$

where

$$\phi(h_j, y_j) = \sum_{l=1}^{p_j} f_l * \delta(v_l, y_j) \quad (5)$$

From a viewpoint of implementation efficiency, formula (4) can be presented in form of (6).

$$d_4(H, Y) = \frac{\sum_{j=1}^{m} \psi(h_j, y_j)}{n} \quad (6)$$

where

$$\psi(h_j, y_j) = \sum_{l=1}^{p_j} f_l * (1 - \delta(v_l, y_j)) \quad (7)$$

Formula (6) can be computed more efficiently for which requires only the frequencies of matched attribute value pairs.

## 3. The $k$-histograms algorithm

The $k$-means algorithm [10], one of the mostly used clustering algorithms, is classified as a *partitional* or *nonhierarchical* clustering method. Given a set of numeric objects $D$ and an integer number $k$ ($\leq n$), the $k$-means algorithm searches for a partition of $D$ into $k$ clusters that minimises the within groups sum of squared errors (WGSS). This process is often formulated as the following mathematical program problem $P$:

$$\text{Minimise } P(W,Q) = \sum_{l=1}^{k}\sum_{i=1}^{n} w_{i,l} d(X_i, Q_l) \tag{8}$$

$$\text{Subject to } \sum_{l=1}^{k} w_{i,l} = 1, \quad 1 \le i \le n$$

$$w_{i,l} \in \{0,1\}, \quad 1 \le i \le n, \quad 1 \le l \le k \tag{9}$$

where $W$ is an $n \times k$ partition matrix, $Q = \{Q_1, Q_2, \ldots, Q_k\}$ is a set of objects in the same object domain, and $d(\cdot,\cdot)$ is the squared Euclidean distance between two objects.

Problem $P$ can be solved by iteratively solving the following two problems:

1. Problem $P1$: Fix $Q = \hat{Q}$ and solve the reduced problem $P(W, \hat{Q})$.

2. Problem $P2$: Fix $W = \hat{W}$ and solve the reduced problem $P(\hat{W}, Q)$.

Problem $P1$ is solved by

$$w_{i,l} = 1 \quad \text{if } d(X_i, Q_l) \le d(X_i, Q_t), \text{ for } 1 \le t \le k \tag{10}$$

$$w_{i,l} = 0 \quad \text{for } 1 \le t \le k$$

and problem $P2$ is solved by

$$q_{l,j} = \frac{\sum_{i=1}^{n} w_{i,l} x_{i,j}}{\sum_{i=1}^{n} w_{i,l}} \tag{11}$$

for $1 \le l \le k$, and $1 \le j \le m$

The basic algorithm to solve problem $P$ is given as follows:

1. Choose an initial $Q^0$ and solve $P(W, Q^0)$ to obtain $W^0$. Set $t = 0$.

2. Let $\hat{W} = W^t$ and solve $P(\hat{W}, Q)$ to obtain $Q^{t+1}$. If $P(\hat{W}, Q^t) = P(\hat{W}, Q^{t+1})$, output $\hat{W}, Q^t$ and stop; otherwise, go to 3.

3. Let $\hat{Q} = Q^{t+1}$ and solve $P(W, \hat{Q})$ to obtain $W^{t+1}$. If $P(W^t, \hat{Q}) = P(W^{t+1}, \hat{Q})$, output $W^t, \hat{Q}$ and stop; otherwise, let $t = t + 1$ and go to 2.

The computational cost of the algorithm is $O(Tkn)$, where $T$ is the number of iterations and $n$

the number of objects in the input data set.

In principle the formulation of problem *P* in the above is also valid for categorical data objects. The reason that the *k*-means algorithm cannot cluster categorical objects is its dissimilarity measure used to solve problem P2. These barriers can be removed by making the following modifications:

1. Replacing means of clusters by histograms.
2. Defining a dissimilarity measure between categorical object and histograms.

Therefore, when (6) is applied as the dissimilarity measure, the cost function (8) becomes

$$P(W,H) = \sum_{l=1}^{k}\sum_{i=1}^{n} w_{i,l} d_4(X_i, H_l) \qquad (12)$$

where $w_{i,l} \in W$ and $H_l = \{h_{l,1}, h_{l,2}, \ldots, h_{l,m}\} \in H$.

To minimize the cost function, the *k*-histogram algorithm uses the dissimilarity measure (6) to solve *P1*, using histograms for clusters instead of means and updating histograms to solve *P2*.

In the *k*-histogram algorithm we need to calculate the total cost *P* against the whole data set each time when a new **H** or *W* is obtained. To make the computation more efficient we use the following algorithm instead in practice that is also adopted in *k*-modes algorithm [7,9].

1. Select *k* initial histograms, one for each cluster.
2. Allocate an object to the cluster whose histogram is the nearest to it according to (6). Update the histogram of the cluster after each allocation.
3. After all objects have been allocated to clusters, retest the dissimilarity of objects against the current histograms. If an object is found such that its nearest histogram belongs to another cluster rather than its current one, reallocate the object to that cluster and update the histograms of both clusters.
4. Repeat 3 until no object has changed clusters after a full cycle test of the whole data set.

Like the *k*-means algorithm the *k*-histograms algorithm also produces locally optimal solutions that are dependent on the initial histograms and the order of objects in the data set. In our current implementation of the *k*-histograms algorithm, our initial histograms selection method selects the first *k* distinct records from the data set to construct initial *k* histograms.

## 4. Experimental Results

A comprehensive performance study has been conducted to evaluate our method. In this section, we describe those experiments and their results. We ran our algorithm on real-life datasets obtained from the UCI Machine Learning Repository [11] to test its clustering performance against other algorithms.

Our algorithms were implemented in Java. All experiments were conducted on a Pentium4−2.4G machine with 512 M of RAM and running Windows 2000.

## 4.1 Real Life Datasets

We experimented with two real-life datasets: the Congressional Voting dataset and the Mushroom dataset, which were obtained from the UCI Machine Learning Repository [11]. Now we will give a brief introduction about these two datasets.

- **Congressional Votes:** It is the United States Congressional Voting Records in 1984. Each tuple represents one Congressman's votes on 16 issues. All attributes are Boolean with Yes (donated as $y$) and No (donated as $n$) values. A classification label of Republican or Democrat is provided with each tuple. The dataset contains 435 tuples with 168 Republicans and 267 Democrats.
- **Mushroom:** The mushroom dataset has 22 attributes with 8124 tuples. Each tuple records physical characteristics of a single mushroom. A classification label of poisonous or edible is provided with each tuple. The numbers of edible and poisonous mushrooms in the dataset are 4208 and 3916, respectively.

In addition, the clustering accuracy for measuring the clustering results is computed as the follows. Suppose the final number of clusters is $k$, clustering accuracy $r$ is defined as: $r = \dfrac{\sum_{i=1}^{k} a_i}{n}$, where $n$ is number of instances in the dataset, $a_i$ is number of instances occurring in both cluster $i$ and its corresponding class, which has the maximal value. In other words, $a_i$ is the number of instances with class label that dominate cluster $i$. Consequently, the clustering error is defined as $e = 1-r$.

## 4.2 Clustering performance

In this section, we compare $k$-histograms with the $k$-modes algorithm [7,9]. The $k$-histograms algorithm is implemented according to the description in this paper, and the $k$-modes algorithm is done strictly according to [9]. To make the comparison fair, both $k$-histograms and the $k$-modes use same initial points selection method, that is, selects the first $k$ distinct records from the data set to construct initial $k$ histograms (or modes).

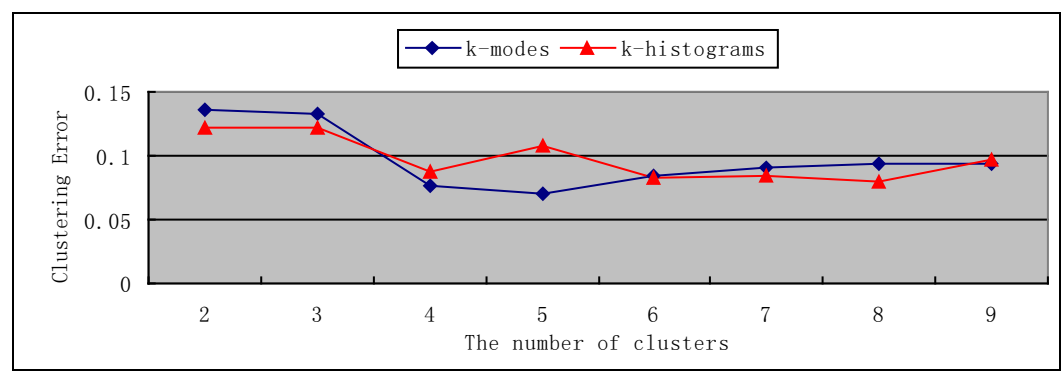

**Fig.1.** Clustering error vs. Different number of clusters (*congressional voting* dataset)

We used the *k*-histograms and the *k*-modes algorithms to cluster the *congressional voting* dataset and the *mushroom* dataset into different numbers of clusters. For each fixed number of clusters, the clustering errors of different algorithms were compared.

Fig.1 contrast the clustering accuracies on the *congressional voting* dataset. From Figure 1, we can summarize the relative performance of our algorithms as follows (See Table 1):

**Table 1:** Relative performance of different clustering algorithms (*congressional voting* dataset)

| Ranking | 1 | 2 |
|---|---|---|
| *k*-modes | 3 | 5 |
| *k*-histograms | 5 | 3 |

That is, comparing to the *k*-modes algorithm, the *k*-histograms algorithm performs the best for 5 cases and the worst in 3 cases. The average clustering accuracy of our algorithm is only a little better than that of the *k*-modes algorithm in this case. However, from this experiment, we are confident to claim that *k*-histograms can provide at least the same level of accuracy as *k*-modes.

Fig.2 contrasts the number of iterations and Fig.3 contrasts the number of objects changed clusters. For the *congressional voting* dataset contains many outliers, from these two figures, we can see that the *k*-histograms algorithm takes less swaps of objects than that of *k*-modes. It indicates that the *k*-histograms algorithm is more robust to outliers than *k*-modes.

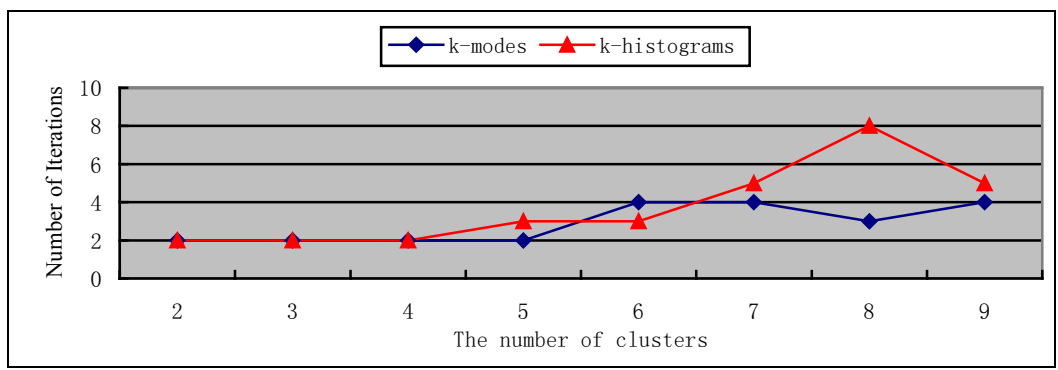

**Fig.2.** Number of iterations vs. Different number of clusters (*congressional voting* dataset)

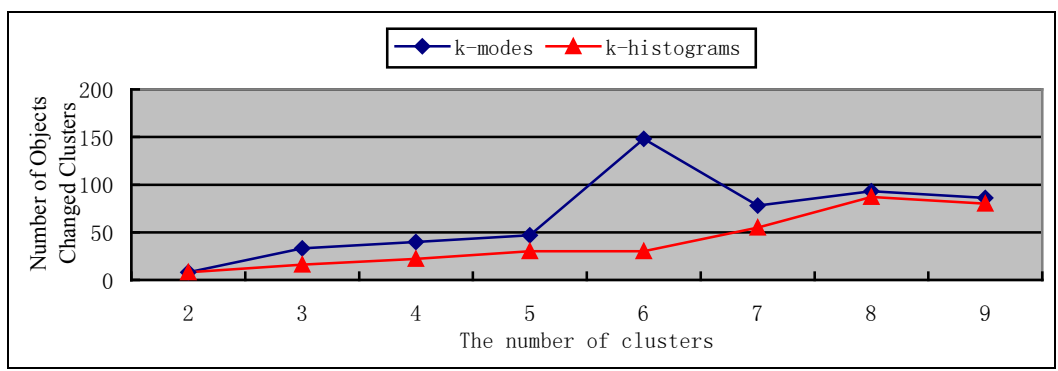

**Fig.3.** Number of objects changed clusters vs. Different number of clusters (*congressional voting* dataset)

The accuracy comparisons on the mushroom dataset are described in Fig. 4 and the summarization on the relative performance of the 2 algorithms is given in Table 2. From these results, we know that *k*-histograms algorithm has better clustering accuracy than *k*-modes. Furthermore, with the increase of the number of clusters, the clustering performance of our algorithm improved consistently, in contrast to the *k*-modes algorithm.

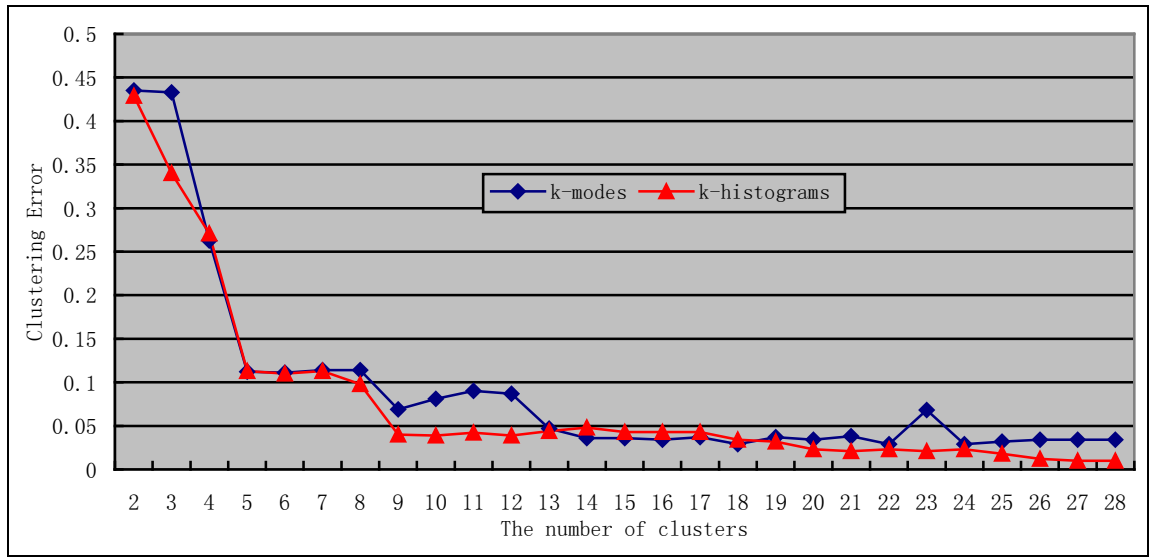

**Fig.4.** Clustering error vs. Different number of clusters (*mushroom* dataset)

**Table 2:** Relative performance of different clustering algorithms (*mushroom* dataset)

| Ranking | 1 | 2 |
|---|---|---|
| *k*-modes | 7 | 19 |
| *k*-histograms | 19 | 7 |

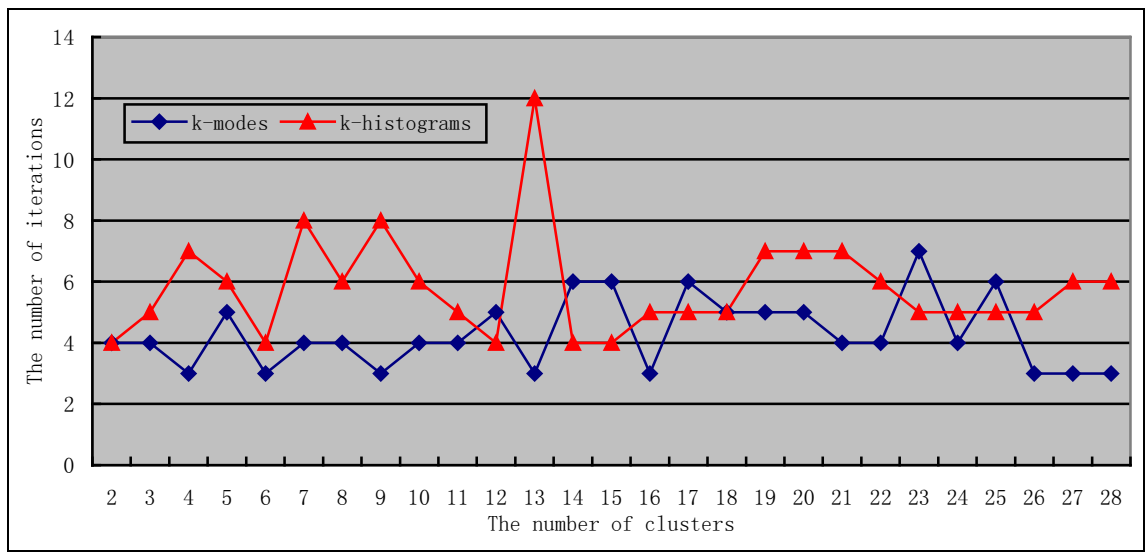

**Fig.5.** Number of iterations vs. Different number of clusters (*mushroom* dataset)

As shown in Fig.5, although the average number of iterations of our algorithm is a little larger than that of the *k*-modes algorithm, while the number of objects changed clusters is much smaller (see Fig. 6), which means the clusters formed in the clustering process of *k*-histograms algorithm are more stable and the algorithm will converge faster.

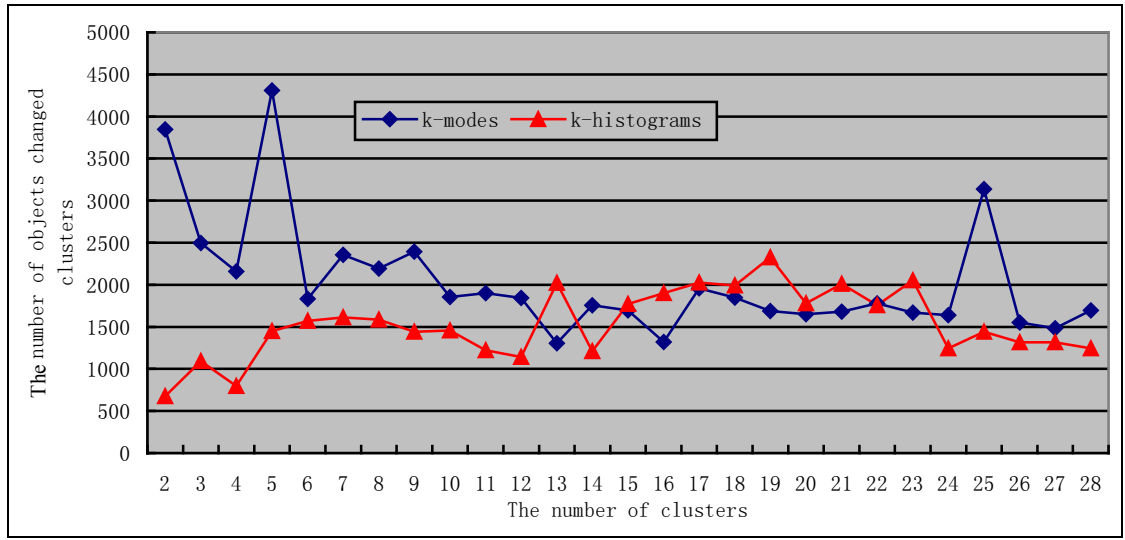

**Fig.6.** Number of objects changed clusters vs. Different number of clusters (*mushroom* dataset)

A cluster is called a pure cluster if all the data objects in it belong to a single class. For the *mushroom* dataset, mushrooms in a "pure" cluster will be either all edible or all poisonous. To find out the ability of the two algorithms on detecting pure clusters, we ran a series of experiments with increasing number of clusters.

Fig.7 presents the comparison on "pure" cluster detection. Compared to the *k*-modes algorithm, our algorithm performed the best in all cases. It never performed the worst. Furthermore, when the number of clusters is increased, the outputs of our algorithm are significantly better than that of the *k*-modes algorithm.

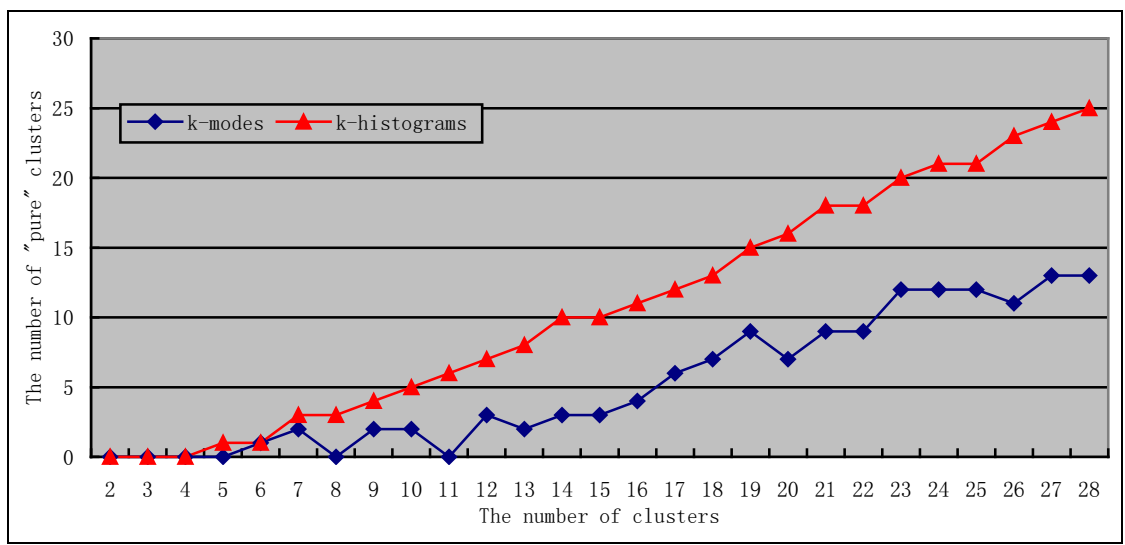

**Fig.7.** Number of "pure" clusters vs. Different number of clusters (*mushroom* dataset)

## 5. Discussions and Conclusions

The *k*-histogram algorithm extends the *k*-means algorithm to categorical domain by replacing the means of clusters with histograms. It is found that the clustering results produced by the proposed algorithm are very high in accuracy.

In general, the *k*-histogram algorithm is very similar to the *k*-modes algorithm. To describe the relationships between these two algorithms, we use attribute value frequency as a basic metric. Moreover, we define *Attribute Value Frequency Threshold* (*AVFT*) that can be differently decided by the individual user. Fig.8 explains the concept of *AVFT* and provides how the *k*-histograms algorithm, *k*-modes algorithm can be distinguished and linked.

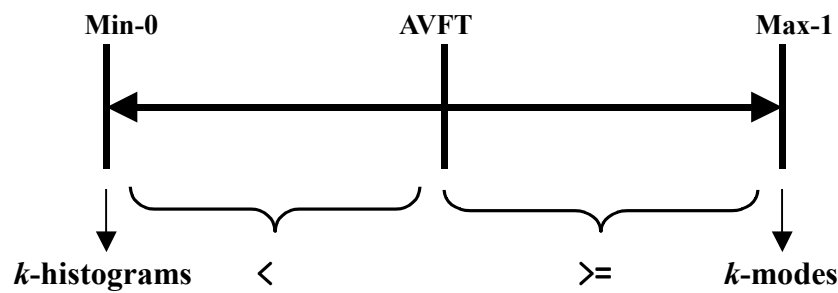

From the viewpoint of *AVFT*, all attribute values are utilized in the *k*-histogram algorithm for representing the cluster (That is, *AVFT* is set to its minimal value, 0), while in *k*-modes algorithm, only the attribute value with maximal frequency in each attribute is used as the representation of the cluster (*AVFT* reach its maximal value). Therefore, in the *AVFT* framework, the *k*-histograms algorithm and *k*-modes algorithm are just two extreme cases in which *AVFT* is set to its extreme values.

From the above discussions, we know that many research issues are still unexplored. For example: Will the algorithms with other *AVFT* values have better clustering performance than that of *k*-histograms and *k*-modes? How to specify a proper *AVFT* value? How to select a proper *AVFT* algorithm with respect to the value distribution of underlying categorical dataset? Are there any criteria that can be followed to do the selection? These questions will be addressed in our future research.